# Learning Shallow Detection Cascades for Wearable Sensor-Based Mobile Health Applications

**Hamid Dadkhahi**  HDADKHAHI@CS.UMASS.EDU
College of Information and Computer Sciences, University of Massachusetts Amherst

**Nazir Saleheen**  NSLEHEEN@MEMPHIS.EDU
**Santosh Kumar**  SKUMAR4@MEMPHIS.EDU
Department of Computer Science, University of Memphis

**Benjamin Marlin**  MARLIN@CS.UMASS.EDU
College of Information and Computer Sciences, University of Massachusetts Amherst

## Abstract

The field of mobile health aims to leverage recent advances in wearable on-body sensing technology and smart phone computing capabilities to develop systems that can monitor health states and deliver just-in-time adaptive interventions. However, existing work has largely focused on analyzing collected data in the off-line setting. In this paper, we propose a novel approach to learning shallow detection cascades developed explicitly for use in a real-time wearable-phone or wearable-phone-cloud systems. We apply our approach to the problem of cigarette smoking detection from a combination of wrist-worn actigraphy data and respiration chest band data using two and three stage cascades.

## 1. Introduction

The field of mobile health or mHealth (Kumar et al., 2013) aims to leverage recent advances in wearable on-body sensing technology and mobile computing to develop systems that can monitor health states and deliver just-in-time adaptive interventions (Nahum-Shani et al., 2014). mHealth research currently targets a wide range of health end points including stress (Plarre et al., 2011), smoking (Ali et al., 2012; Saleheen et al., 2015), eating (Thomaz et al., 2015), and even drug use (Natarajan et al., 2013; Hossain et al., 2014).

However, mHealth research using wearable sensors has focused almost exclusively on passive data collection followed by offline data analysis based on common machine learning models and algorithms including support vector machines (Cortes & Vapnik, 1995) and random forests (Breiman, 2001). While this is an important first step, much current mHealth research on detection models implicitly assumes that features from all sensors are available simultaneously, that compute resources are unbounded, and that results do not need to be delivered in real time. These assumptions are clearly not valid for tasks including real-time monitoring, and just-in-time-adaptive interventions. These assumptions also discount the fact that such applications must be deployed within a heterogeneous, resource constrained wearable-phone or wearable-phone-cloud environment, which necessitates minimizing energy use and communication costs.

For example, the applications mentioned earlier use one or more wearable sensing devices including smart watches like the Microsoft Band or Pebble watch, and chest band sensors like the Zephyr BioHarness. These devices typically have limited energy and compute resources due to their small form factors. The wearable sensors are linked with a smart phone (typically using Bluetooth) that has greater, but still limited, energy and compute resources. The smart phone can in turn have access to cloud-based compute resources (typically using WiFi or cellular networks). However, communication across devices and communication with the cloud requires both time and energy.

In this paper we take a first step towards addressing these problems by developing a novel approach to learning shallow (two or three-stage) cascades of heterogeneous detection models that are designed to be deployed on a heterogeneous device cascade. Unlike prior work on learning cascades based on boosting (Freund & Schapire, 1997) and the soft cascade learning framework (Raykar et al., 2010), we train all stages jointly using an objective function that more closely mimics the application of the models in the hard decision setting, while simultaneously minimizing the cost incurred when data cases pass between stages. We also match cascade stages to the resources available on the corresponding physical compute devices.

We present experiments comparing our approach to single models as well as the soft cascade framework using data from the smoking detection domain. This data set includes sensor data streams from both a wrist-worn actigraphy sensor and a respiration chest band sensor. Initial results for two and three-stage cascades show that our approach can achieve a substantial reduction in computation time relative to using a single complex model, while achieving a better speed-accuracy trade-off than the soft cascade framework when applied to the same cascade architecture.

## 2. Related Work

A classifier cascade is a collection of models that are applied in sequence to classify a data instance. In order for a data instance to be classified as positive, it must be classified as positive by all stages in the cascade. If any stage in the cascade rejects a data instance, processing of that instance immediately stops and it is classified as a negative instance. For highly class-imbalanced data, cascades can lead to substantial computational speedups.

Perhaps the most well-known work on classifier cascade learning is the Viola-Jones face detection framework (Viola & Jones, 2001). This framework trains a classification model for each stage sequentially using a boosting algorithm (Freund & Schapire, 1997). Each stage is trained by boosting single-feature threshold classifiers by training only on the positive examples propagated by the previous stage. The bias of the final boosted model for each stage is then adjusted to minimize the number of false negatives. The Viola-Jones cascade achieves real-time face detection by quickly rejecting the vast majority of sub-windows in an image that do not contain a face.

Subsequent work on boosting-based learning for cascades has focused on a number of shortcomings of the Viola-Jones cascade including extensions of adaboost for improved design of the cascade stages, joint training instead of greedy stage-wise training, and methods for learning optimal configurations of a boosted cascade including the number of boosting rounds per stage and the number of total stages. Saberian et al. present an excellent discussion of this work (Saberian & Vasconcelos, 2014).

An alternative to boosting for cascade learning is the noisy-AND approach (Lefakis & Fleuret, 2010). In this framework, the probability that an instance is classified as positive is given by the product of the output probabilities of an ensemble of base classifiers (often logistic regression models). If any element of the ensemble predicts a negative label for a data instance, the instance will receive a negative label. The models in the ensemble are trained jointly using the cross-entropy loss applied to the product of their probabilities. For deployment as a cascade, the learned models can be placed in a sequence.

A disadvantage of the noisy-AND approach is that there is no explicit penalization related to how many stages a data case propagates through before it is rejected as a negative example. Raykar et al. proposed a modification to the noisy-AND approach that retained the cross-entropy/noisy-AND objective, but added a penalty term to penalize the model based on the number of stages required to reject an example (Raykar et al., 2010). They refer to their approach as a "soft cascade." The primary disadvantage of their approach is that the cascade is still operated using hard decisions, which is not well-matched to the training objective. Our approach is closest to that of Raykar et al., but uses a shallow cascade of heterogeneous models trained using a primary objective function that better approximates the hard decisions that occur in an actual deployed cascade.

## 3. Proposed Cascaded Model

To begin, assume we wish to learn a cascaded model consisting of $L$ stages. We define a probabilistic classifier $P_l(y|\mathbf{x})$ for each stage $l$. We let the output of the cascade be $P_*(y|\mathbf{x})$. In the noisy-AND and soft cascade frameworks introduced in the previous section, $P_*(y|\mathbf{x})$ is defined as shown below:

$$P_*(y|\mathbf{x}) = \prod_{l=1}^{L} P_l(y|\mathbf{x}) \qquad (1)$$

We propose an alternative combination rule that better reflects the idea that in a hard cascade the output of each stage of the cascade gates the computation of the subsequent stage. Our combination rule for a general cascade is given below. We use the shorthand $P_l(y|\mathbf{x}) = p_l$ to simplify the notation.

$$P_*(y|\mathbf{x}) = \sum_{l=1}^{L} \theta_l \cdot P_l \qquad (2)$$

$$\theta_l = \begin{cases} \left(1 - g_\alpha(P_l)\right) \prod_{k=1}^{l-1} g_\alpha(P_k) & l < L \\ \prod_{k=1}^{L-1} g_\alpha(P_k) & l = L \end{cases} \qquad (3)$$

$$g_\alpha(p) = \frac{1}{1 + \exp(-\alpha(p - 0.5))} \qquad (4)$$

Equations 2 to 4 show that our proposed model takes the form of a mixture of experts (Jordan & Jacobs, 1994) with highly specialized mixture weights. The effect of these mixture weights is to place nearly all of the weight in the mixture either on the output of the first stage in the cascade that classifies an instance as negative, or on the output of the classifier in the last stage of the cascade. This is accomplished using the logistic function $g_\alpha(p)$ shown in Equation 4 with a large value of $\alpha$ to approximate the hard decision rule actually used in the deployed cascade.

In Equation 5, we give an example of the explicit form of a three-stage cascade to further clarify the cascade design.

$$P_*(y|\mathbf{x}) = (1 - g_\alpha(p_1)) \cdot p_1 + g_\alpha(p_1)(1 - g_\alpha(p_2)) \cdot p_2 \\ + g_\alpha(p_1)g_\alpha(p_2) \cdot p_3 \quad (5)$$

With a large value of $\alpha$ (we use $\alpha = 100$), $g_\alpha(p)$ approximates a step function with the step located at $0.5$. If the output of the first stage $p_1$ is somewhat less than $0.5$, $g_\alpha(p_1)$ will be approximately zero and the output of the cascade will be $P_*(y|\mathbf{x}) \approx p_1$. If the output of the first stage is greater than $0.5$, but the output of the second stage is less than $0.5$ then $g_\alpha(p_1)$ will be approximately 1 while $g_\alpha(p_2)$ will be approximately 0 and the output of the cascade will be $P_*(y|\mathbf{x}) \approx p_2$. Finally, if both $p_1$ and $p_2$ are greater than $0.5$, then both $g_\alpha(p_1)$ and $g_\alpha(p_2)$ will be approximately 1 and the output of the cascade will be $P_*(y|\mathbf{x}) \approx p_3$. Thus, the probability output by the cascade will be approximately equal to either the output of the first stage $l$ to reject a data instance with $p_l < 0.5$, or the output of the final stage, $p_L$. Interestingly, the model can thus be viewed as a self-gated mixture of experts since the usual gating function is replaced by a gating function based on the outputs of the experts themselves.

To learn the model, we maximize the log likelihood of $P_*(y|\mathbf{x})$ (equivalent to minimizing the cross entropy loss), subject to a per-instance regularizer $r(y_n, \mathbf{x}_n)$. The objective function is shown below where we let the data set $\mathcal{D} = \{(y_n, \mathbf{x}_n) | 1 \leq n \leq N\}$ and $N$ is the number of data instances.

$$\mathcal{L}(\mathcal{D}) = \sum_{n=1}^{N} \ell(y_n, \mathbf{x}_n) + \lambda r(y_n, \mathbf{x}_n) \quad (6)$$

$$\ell(y, \mathbf{x}) = y \log P_*(y|\mathbf{x}) + (1 - y) \log(1 - P_*(y|\mathbf{x})) \quad (7)$$

$$r(y, \mathbf{x}) = \kappa_1 + \sum_{l=2}^{L} \kappa_l \prod_{k=1}^{(l-1)} g_\alpha(P_k(y|\mathbf{x})) \quad (8)$$

Again, with a large value of $\alpha$, $g_\alpha(P_l(y|\mathbf{x}))$ will be approximately 0 for stages that output values that are less than $0.5$, and will be approximately 1 for stages that are greater than $0.5$. Thus, this regularizer applies a penalty approximately equal to the total cost of executing the number of stages actually used in the cascade to classify a given instance, where $\kappa_l$ is the cost per stage. It is similar to the penalty function used in (Raykar et al., 2010), but is a better match to a hard cascade due to approximating the step function with the $g_\alpha()$ function.

As mentioned in the introduction, our interest in this work is the application of this model to the case of shallow cascades corresponding to a wearable-phone or wearable-phone-cloud system architecture. Unlike most earlier work on boosted cascades, there is a direct mapping between the features available at a given stage and the hardware that stage runs on, so there is much more limited flexibility in the assignment of features to stages. In addition, the hardware that the stages run on becomes increasingly more powerful down the cascade, so it is natural to consider using a cascade of heterogeneous classifiers. In this work, we consider the use of a logistic regression classifier in the first stage, followed by the application of neural network based classifiers in subsequent stages.

The complete set of models used in a given cascade can be optimized jointly using the objective function described above. In addition, since the models used in later stages of the cascade are increasingly powerful, we can also initialize training by learning the models in reverse order from layer $L$ to layer 1, with the model for layer $l$ being able to depend on the downstream performance of layers $l + 1$ to $L$. We use this initialization combined with fine tuning the cascade using joint training in the experiments that follow. The precise cascade architectures we consider in this work are described in the next section.

## 4. Experiments and Results

In this section, we present experimental results comparing our proposed cascade architecture to the soft cascade of (Raykar et al., 2010). As a test bed, we use the PuffMarker smoking detection dataset from (Saleheen et al., 2015). In the PuffMarker data set, each data case consists of 37 features. 19 features are computed from a respiratory inductance plethysmography sensor data stream, and 13 features are computed from accelerometer and gyroscope sensors on a wrist band. Overall, there are 3836 data cases in the PuffMarker dataset. We consider a stratified division of the data into 3400 training cases (with 260 cases in the positive class) and 436 test cases (with 31 cases in the positive class).

We compare a single-stage model to two and three-stage cascades trained using both our proposed approach and the soft cascade approach. For a single-stage model, we use a one-hidden-layer neural network (1LNN) with $K = 10$ hidden units and all 37 features. In all cascade models, we use logistic regression (LR) in the first stage. We consider 5 features (after z-score normalization) obtained via the basis expansion $\Phi : [x, y] \to [x, y, x^2, y^2, xy]$ applied to roll ($x$) and pitch ($y$) features computed from the accelerometer data stream. This transformation is suggested by Figure 7 in Saleheen et al. (2015). For comparison, we also train models using all 37 features in the first stage.

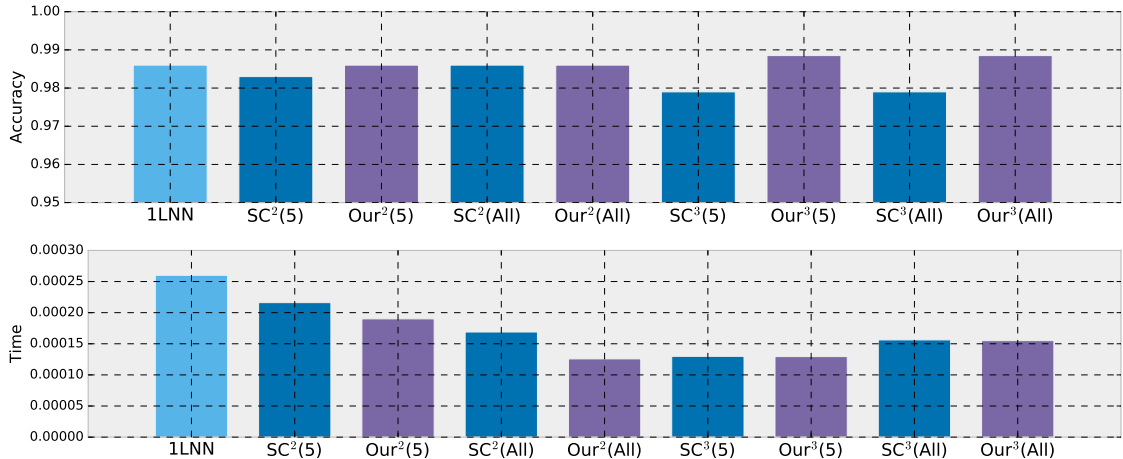

*Figure 1.* Evaluation of different cascade models in terms of accuracy (top) and classification time (bottom). `SC` and `Our` correspond to the soft cascade model and our proposed cascade model, respectively. In both cases, the superscript indicates the number of stages in the cascade model, and the numbers inside the parentheses indicates the number of features used in the first stage of the cascade.

For two-stage models, we use a one-hidden-layer neural network (1LNN) with $K = 10$ hidden units. For three-stage models, we use a one-layer neural network (1LNN) with $K_1 = 3$ hidden units as the second-stage classifier and a two-layer neural network (2LNN) with $K_1 = 10$ and $K_2 = 20$ hidden units as the third-stage classifier. All models in the second and third stages use logistic non-linearities and all 37 features. Preliminary testing was used to identify the hidden layer sizes. Using larger hidden layer sizes tends to either result in lower accuracy due to over-fitting or similar accuracy, but increased time.

We assume a cost-per stage that is proportional to the compute time for that stage. The regularization parameters in our proposed cascade model and the soft cascade model were swept over a grid to produce a speed-accuracy trade-off curve (or surface). We compare approaches by identifying the maximum accuracy setting of the regularization parameters for each cascade architecture, and then compare the time that the methods require to achieve that accuracy. All experiments were performed on 2.4GHz Intel Xeon E5-2440 CPU's. Timing results are averaged over 10,000 classifier evaluations.

The results are shown in Figure 1. First, we can see that all of the cascaded models outperform the single-stage classifier in terms of classification time. Our proposed approach also obtains the same or lower classification time compared to the soft cascade model for every cascade architecture considered. We can also see that in all of the cases where our approach obtains the same classification time as the soft cascade, it does so while achieving higher accuracy. We note that our proposed three-stage model is actually able to outperform the single stage model in terms of accuracy while requiring approximately half the time. We note that similar accuracy can be obtained using a single-stage three-layer neural network model, but our model takes one quarter the time of this single-stage three layer model. Finally, we note that the maximum accuracies that we obtain are similar to those in Saleheen et al. (2015) (98.7%).

## 5. Conclusions and Future Work

We have introduced a new approach to cascaded classifier learning using a cascade architecture that better matches the hard decisions that are made when the cascade is applied at detection time. Our initial results are promising and show that our proposed cascade architecture outperforms the soft cascade architecture in terms of a speed-accuracy trade-off.

Future work will address several important limitations of this study. First, better cost models need to be developed for the devices that we intend to deploy cascades on. The current study uses computation time as proxy, but real applications need to consider a more general energy-based cost model that takes into consideration the cost of sensing, computing, and communicating across devices. Second, we intend to deploy the learned smoking detection models on actual hardware to assess the performance of the end-to-end system. Finally, we plan to expand the application of the proposed architecture to other application domains and other model types. Of particular interest are more intensive structured prediction-based detection models (for example, conditional random field models) where cloud-based computation is likely to be required.

# 6. Acknowledgments

The authors would like to thank Deepak Ganesan for helpful discussions of this research. This work was partially supported by the National Institutes of Health under awards R01DA033733, R01DA035502, 1R01CA190329, R01MD010362, and 1U54EB020404, and the National Science Foundation under awards IIS-1350522 and IIS-1231754.